\pdfoutput=1

\documentclass[11pt]{article}
\usepackage[]{acl}

\usepackage{cuted}%
\usepackage[]{amsthm}
\usepackage{times}
\usepackage{latexsym}

\usepackage{adjustbox}
\usepackage{graphicx}

\usepackage[T1]{fontenc}
\graphicspath{ {./images/} } 
\newcommand{\ignore}[1]{}

\usepackage[utf8]{inputenc}

\usepackage{microtype}

\usepackage{booktabs}
\usepackage{multirow}
\usepackage{graphicx}
\usepackage[normalem]{ulem}
\useunder{\uline}{\ul}{}

\title{iLab at SemEval-2023 Task 11 Le-Wi-Di: \\ Modelling
Disagreement or Modelling Perspectives?}

\author{Nikolas Vitsakis$^{1}$ \and Amit Parekh$^{1}$ \and Tanvi Dinkar$^{1}$ \\ \and \textbf{Gavin Abercrombie}$^{1}$ \and \textbf{Ioannis Konstas}$^{1,2}$ \and \textbf{Verena Rieser}$^{1,2}$ \\
        $^{1}$The Interaction Lab, Heriot-Watt University \ \
        $^{2}$Alana AI \\
        \texttt{nv2006@hw.ac.uk}}

\def\GAdel#1{\bgroup\markoverwith{\textcolor{blue}{\rule[0.5ex]{2pt}{1pt}}}\ULon{#1}}

\def\TDdel#1{\bgroup\markoverwith{\textcolor{teal}{\rule[0.5ex]{2pt}{1pt}}}\ULon{#1}}

\def\NVdel#1{\bgroup\markoverwith{\textcolor{magenta}{\rule[0.5ex]{2pt}{1pt}}}\ULon{#1}}

\begin{document}
\maketitle
\begin{abstract}
There are two competing approaches for modelling annotator disagreement: distributional soft-labelling approaches (which aim to capture the level of disagreement) or modelling perspectives of individual annotators or groups thereof. We adapt a multi-task architecture ---which has previously shown success in modelling perspectives--- to evaluate its performance on the SEMEVAL Task 11. We do so by combining both approaches, i.e. predicting individual annotator perspectives as an \emph{interim step} towards predicting annotator disagreement. Despite its previous success, we found that a multi-task approach performed poorly on datasets which contained distinct annotator opinions, suggesting that this approach may not always be suitable when modelling perspectives. Furthermore, our results explain that while strongly perspectivist approaches might not achieve state-of-the-art performance according to evaluation metrics used by distributional approaches, our approach allows for a more nuanced understanding of individual perspectives present in the data. We argue that perspectivist approaches are preferable because they enable decision makers to amplify minority views, and that it is important to re-evaluate metrics to reflect this goal.

\ignore{   SEMEVAL Task 11 encourages participating teams to produce solutions to the problem of subjective classification tasks in scenarios in which annotators do not agree.
    We propose a multi-task learning approach to the classification of abusive language in the four shared-task datasets.
    Our approach is an example of \emph{strong perspectivism}, in which we aim to model not only the level of disagreement between labels, but also the perspectives of the individual annotators.
    We find that while \ldots }
\end{abstract}

\section{Introduction}

\ignore{There has been 
growing interest in  applying natural language processing methods to social computing tasks, such as toxicity and hate speech detection~\citep[e.g.][]{vidgen2020directions}. Often this involves training models via a supervised learning paradigm; i.e. classification of labelled data.} Many Natural Language Processing (NLP) tasks follow a supervised learning paradigm, i.e. classification of labelled data where multiple annotations are aggregated
into a \emph{hard label} using averaging or majority voting. Hard labels are based on the assumption that each instance in a dataset has \emph{one singularly correct response}---often referred to as `ground truth'. 

However, this assumption is \emph{highly unrealistic} for social computing tasks, such as such as toxicity and hate speech detection~\citep[e.g.][]{vidgen2020directions}, where lived experiences, biases, opinions and annotator experience all play a role in the \emph{subjective} response an annotator might give. Hard labels especially disadvantage minority groups \cite{blodgett-2021-socio}. For example, in abusive language classification, where a minority is disproportionately affected (such as minoritised people who have faced online harassment), an aggregated majority label can obscure the perspective of the most vulnerable groups. 

Thus, there is growing awareness that modelling multiple perspectives is necessary, particularly for inherently subjective tasks and those concerned with social issues \citep{nlperspectives-2022-perspectivist,cabitza-etal-2023-toward,plank-2022-problem}. 

\noindent \paragraph{Le-Wi-Di}
The SemEval 2023 shared task `Learning With Disagreements'~\citep{LeWiDi2023semeval} aims to capture and model annotator disagreement -- going beyond the assumption of one aggregated `ground truth' label. Participating teams are required to propose methods that consider \emph{disaggregated annotations}, used to create a \emph{soft label}, which represents the probabilistic distribution of the annotations. Soft labels can then be used to predict the \emph{level of disagreement} for each instance in a dataset. 

The task presents a benchmark of four datasets, including a hard label and soft labels for each instance. The datasets were chosen specifically to represent tasks that are highly subjective (e.g. hate speech detection) and show high annotator disagreement. Participating teams are evaluated on how well their proposed model predicts both hard and soft labels, via F1 accuracy score and cross-entropy loss, respectively. The shared task prioritised the soft evaluation, i.e how well the model's probabilities reflect annotator agreement, rather than simply proposing models that would outperform the current state-of-the-art for the hard labels. 
For further details of the datasets, see \autoref{sec: data}.

\paragraph{Our approach}
proposes a modified version of the multi-task model introduced by~\citet{davani-etal-2022-dealing}, which aims to predict individual annotator judgments. By training our model to predict each annotator's judgement for each instance in a dataset, we can use the resulting predictions to infer the level of disagreement for that instance, without directly training for such a purpose. 

The main benefit of our approach is that opinions present in the dataset are preserved beyond the simplistic form of a polarised agreement/disagreement distribution. Instead, we focus on representing individual opinions, also knows as `perspectives' \cite{cabitza-etal-2023-toward}.This allows modelling of specific perspectives present in a dataset, potentially enabling the amplification of minority opinions. 

\begin{table*}[ht]
 \resizebox{\textwidth}{!}{
\begin{tabular}{@{}cc|cccc@{}}
                                                                              &         & \textbf{HS-Brexit} & \textbf{ArMIS} & \textbf{ConvAbuse} & \textbf{MD} \\ \midrule
Task                              &                       & Hate speech    & Misogyny & Abusiveness    & Offensiveness  \\
\multirow{3}{*}{No. of instances} & train                 & 784            & 657      & 2398           & 6592           \\
                                  & dev                   & 168            & 141      & 812            & 1104           \\
                                  & test                  & 168            & 145      & 840            & 3057           \\ 
Utterance length                  &                       & $18.623\pm4.578$ &   $19.510\pm12.042$     & $27.322\pm18.830$ & $22.614\pm14.777$ \\ \midrule
\multirow{4}{*}{\begin{tabular}[c]{@{}c@{}}Annotator \\ details\end{tabular}} & Krippendorff's $\alpha$ ($\downarrow$) & 0.347              & 0.524          & 0.650              & \textbf{0.359}       \\
                                  & Total annotators      & 6              & 3        & 8              & 819            \\
                                  & Annotators / instance & 6              & 3        & 4              & 5              \\
                                  & Unseen annotators     & 0               &  0        &    0            & \textbf{91}              
\end{tabular}
}
\caption{Descriptive data and annotator statistics: utterance length in tokens with standard deviations, inter-annotator agreement measured with Krippendorff's $\alpha$ (($\downarrow$) lower=higher disagreement), and `unseen annotators', the percentage of annotators that are not represented by at least one instance in all of the train, dev and test sets.}
\label{tab:kripps_score}
\end{table*} 

\section{Related work}

\subsection{Modelling disagreement} 
\citet{uma-etal-2021-learning} provide an extensive survey, outlining four main approaches. The first aggregates annotations into hard labels (as in Task metric 1). The second removes items that display high disagreement and is thus unsuitable for this task. The third models the distribution of annotations for each item, i.e  `soft labels' (as in metric 2). The final approach enables the model to optimise across different tasks through the use of either the use of multi-task learning, or a procedure called plank-style weighing (for more information towards this method refer to \citet{Plank2014LearningPT}). 

For the purposes of this paper, the former is relevant, as multi-task learning, enables the resulting model to optimise across different tasks through the use of both hard and soft labels, providing predictions for the hard label, as well as degrees of confidence for each~\citep{fornaciari-etal-2022-hard}. While these approaches make use of disagreement to enhance optimisation, they have not been used to preserve the different perspectives represented in the data.

\subsection{Modelling perspectives}

We focus on a `strong perspectvist approach' which aims to \emph{preserve diversity of perspectives} throughout the whole classification pipeline \citep{cabitza-etal-2023-toward}. To contrast, weak perspectivist approaches \citep{cabitza-etal-2023-toward} may consider several annotator viewpoints, but still reduce these viewpoints towards a single label. An example of a strong perspectivist approach is \citet{davani-etal-2022-dealing}, who predict individual annotator judgments using a multi-task approach, and treat each annotator as a sub-task, thus retaining individual perspectives. While research has shown some success when utilising single-task models to accurately capture distinct perspectives \cite{rodrigues2018deep}, our choice of model was informed by recent evidence that multi-task models \cite{fornaciari-etal-2021-beyond} can outperform single-task models.

However, one limitation (with respect to strong perspectivism) is that, for evaluation, they aggregate predicted annotations into one label, essentially falling back into the issues of hard labels. Our proposed solution aims to address this limitation through the use of both hard and soft labels, i.e. evaluating model performance on the disaggregated perspectives present in the dataset.

\section{Data}
\label{sec: data}
The Le-Wi-Di\footnote{\url{https://le-wi-di.github.io/}} shared task consists of the following four datasets, that have all been synthesised into a common \texttt{json} format.\\
\textbf{HS-Brexit} \citep{akhtar2021whose}: a dataset of English tweets on the topic of Brexit, labelled for hate speech by six annotators belonging to two different groups; three Muslim immigrants and three other individuals.\\
\textbf{ArMIS} \citep{almanea-poesio-2022-armis}: Arabic tweets annotated for misogyny and sexism. 
The three annotators for this dataset have different self-identified demographics of `Moderate Female', `Liberal Female' and `Conservative Male'.  \\
\textbf{ConvAbuse} \citep{cercas-curry-etal-2021-convabuse}: a dataset of English dialogues between users and two conversational agents. The dialogues have been been annotated for abusiveness by eight gender studies experts.\\
\textbf{MultiDomain Agreement (MD)}: \citep{leonardelli-etal-2021-agreeing}: English tweets on three topics: BLM, Election and Covid-19, labelled for offensiveness by crowdsourced annotators and specifically selected to elicit disagreement. 

See \autoref{tab:kripps_score} for descriptive statistics of the datasets. Disagreement is moderate for the ArMIS and ConvAbuse datasets\footnote{Note, our results for calculating inter-annotator agreement may slightly differ from the original papers (e.g. \citet{cercas-curry-etal-2021-convabuse} Krippendorff's $\alpha \approx 0.69$) given discarded instances etc. Krippendorff's $\alpha$ is calculated using the \href{https://pypi.org/project/krippendorff/}{Fast Krippendorff} package. The values range from $0-1$, where 0 depicts perfect disagreement, and 1 complete agreement.}. While the MD dataset has (unsurprisingly) higher disagreement given the large pool of annotators (891 annotators total), the percentage of unseen annotators (i.e. annotators that don't appear in all splits) is extremely high (91\%) compared to the other datasets (0\%). 
These factors could lead to sparsity of strongly distinguished perspectives present in the MD dataset. In ConvAbuse, the standard deviation of utterance lengths\footnote{Calculated using the \href{https://spacy.io/api/tokeniser/}{Spacy tokeniser} package for English, and \href{https://github.com/aub-mind/arabert}{Arabert} package for Arabic.} is higher due to the presence of many single token responses that can be present in a dialogue---such as `\emph{yes}'. 

\begin{figure*}[!ht]
    \centering
    \includegraphics[scale=0.23]{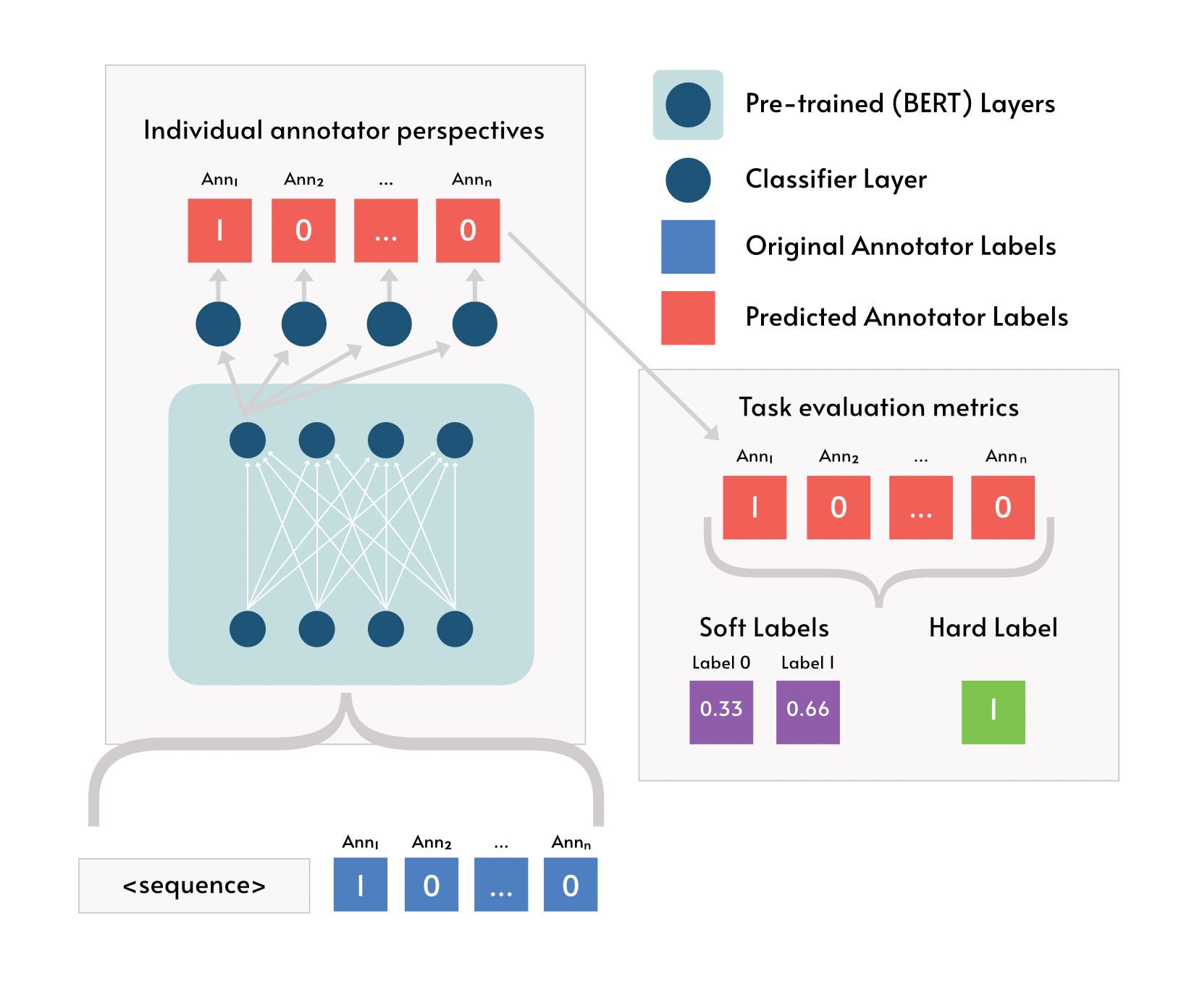}
    \caption{Representation of our multi-task architecture. As shown, we predict individual annotator perspectives (individual cross entropy, shown on the left of the figure) as an interim step to predicting the level of disagreement (the task metric of soft-label cross entropy, as shown on the right). For a full system description, refer to \autoref{sec:system}}
    \label{fig:model1}
\end{figure*}

\section{Methods} \label{sec: methods}

\subsection{System overview}
\label{sec:system}

We implement a multi-task model, which makes, for each instance, separate predictions for each annotator present in the dataset. Given the varied characteristics of our datasets, such as missing labels or large number of annotators, this approach allows for the evaluation of multi-task learning across a variety contexts. An overview of the model and the predicted output is shown in Figure \ref{fig:model1}.

For a given text sample in a dataset, $\mathbf{x} \in \mathbf{X}$, our model $p_\theta(\mathbf{y} | \mathbf{x})$ predicts the individual annotation of each annotator $\mathbf{y} = (y_1, \dots, y_K)$, where $K$ is the total number of unique annotators within the dataset. 
The predicted hard label of an instance is defined as aggregation of predictions into one label $mode(\mathbf{y}) = z$, where $z \in \{0,1\}$, while soft labels as $v_0 \in [0,1]$ and $v_1 \in [0,1]$ that denote the possible probabilistic distributions of the annotations. 

For the purposes of the shared task, we evaluate our models through F1 and cross entropy scores for the hard and soft labels respectively. 
We refer to the cross entropy loss of the shared task as \emph{soft-label cross entropy}.We needed a different way to model our strong perspectivist approach, as the shared task metrics prioritise predicting the \emph{level of disagreement}. 
As seen in \autoref{fig:model1}, our model treated each annotator per instance as a subtask (shown as a classification layer). 
We evaluate the predicted versus the true labels for each annotator using cross entropy loss, which we refer to as \emph{individual cross entropy}, which is also the metric used to train our model. Optimising individual cross entropy should lead to more accurate individual predictions, resulting in a representative annotation matrix, which can, in turn, be used to calculate the soft-label cross entropy.
Since the aim of the shared task was to capture disagreement, we prioritised minimising soft-label cross entropy (CE) loss scores over high performing F1 scores. 

This was done by manually stopping the model's training procedure (for individual cross entropy loss) when the minimum soft-label cross entropy scores were achieved for each dataset. Thus, we do not optimize our model directly using the shared task evaluation metrics. However, these evaluation metrics can still be used as an indirect measure of our model's performance. Hence, we report the shared task metrics on our model, which was trained on minimising the individual cross entropy loss in order to capture perspectives. While this method is not optimised to predict disagreement when compared to models trained by minimising soft-label cross entropy scores, it allows for the prediction of individual annotator perspectives. Thus, we model individual perspectives as an interim step towards predicting disagreement. 

We compare the performance and analyse the benefits of using a multi-task model against the organisers' baseline (aggregated labels), and two other models: a baseline neural model, and an SVM model (further described in \autoref{subsec:models}). The SVM model was used as a linear model baseline, due to its prior success compared to neural approaches for cases of abuse detection~\citep{niemann-etal-2020-abusive}. 

\subsection{Text encoding}

We applied the following pre-processing steps. 
For the ConvAbuse dataset, we only processed the last sentence uttered by a user, as \citet{cercas-curry-etal-2021-convabuse} reported no significant performance improvements from adding dialogue context. We preprocessed the Arabic dataset following \citet{antoun-etal-2020-arabert}. For the SVM model, both English and Arabic datasets were tokenised using term frequency-inverse document frequency (TF-IDF).

\subsection{Model architectures} \label{subsec:models}

\paragraph{Baseline Linear model } 

We trained an SVM model to perform binary classification with a linear kernel using a bag-of-words and TF-IDF approach. 

The model outputs a distribution over the possible hard and soft labels, over which F1 and cross entropy scores were calculated.

\paragraph{Single-task (baseline) BERT Model} 
For our transformer-based models, we used the pre-trained \texttt{BERT-base-uncased} \citep{devlin-etal-2019-bert} model from the HuggingFace Transformers library \citep{wolf-etal-2020-transformers} for initialising English models. For the Arabic dataset, Transformer-based models were initialised with \texttt{AraBERT}, a variant of BERT pre-trained specifically for Arabic text which has shown comparable results to multilingual BERT in NLP-related tasks such as sentiment analysis and Named Entity Recognition \citep{antoun-etal-2020-arabert}. Outputs of both models were fed through a linear layer, with a softmax activation function, resulting in a probability distribution for a binary label. The model used ADAM optimisation \citep{kingma2014adam}. For specific parameters see \autoref{sec:appendix}. Like the SVM model, the model outputs a distribution over the possible hard and soft labels, over which F1 and cross entropy scores were calculated. 

\paragraph{Multi-task BERT Model} 

\begin{table*}[ht!]
    \centering
    \begin{tabular}{l|c|c|c|c|c|c|c|c}
         & \multicolumn{2}{c|}{\textbf{HS-Brexit}} & \multicolumn{2}{c|}{\textbf{ArMIS}} & \multicolumn{2}{c|}{\textbf{ConvAbuse}} & \multicolumn{2}{c}{\textbf{MD}}  \\
         \textbf{Models} & \textbf{F1 ($\uparrow$)} & \textbf{CE ($\downarrow$)} & \textbf{F1 ($\uparrow$)} & \textbf{CE ($\downarrow$)} & \textbf{F1 ($\uparrow$)} & \textbf{CE ($\downarrow$)} & \textbf{F1 ($\uparrow$)} & \textbf{CE ($\downarrow$)} \\
         \hline 
         Baseline (agg.)  & \textbf{0.89} & 2.71 & 0.59 & 8.23 & \textbf{0.95} & 3.38 & 0.78 & 7.74\\ 
         TFIDF - SVM & 0.86 & 0.62 & 0.68 & 2.57 & 0.9 & 0.49 & \textbf{0.88} & 0.62 \\
         Single-task BERT   & 0.47 & \textbf{0.43} & \textbf{0.78} & \textbf{1.77} & 0.88 & \textbf{0.37} & 0.74 & \textbf{0.61} \\
         Multi-Task BERT  & 0.46 & 0.76 & 0.58 & 1.89 & 0.77 & 0.50 & 0.41 & 1.28 \\
    \end{tabular}
    \caption{Model performance using F1 and cross-entropy (CE) scores for the four datasets of SemEval-2023 Task 11 Le-Wi-Di. \textbf{$\uparrow$}, and \textbf{$\downarrow$} indicate that higher and lower scores represent better performance, respectively. Baseline provided by the organisers (aggregation). We highlight in bold the highest scores for each dataset respectively.}
    \label{tab:all_results}
\end{table*}

Our model is a modified version of that proposed by \citet{davani-etal-2022-dealing} as shown in \autoref{fig:model1} and \autoref{sec:system}. A pre-trained BERT model is used to encode the text, upon which we train separate classification layers for each annotator. Training parameters are identical to our baseline transformer model (\autoref{sec:appendix}). The embedded \texttt{[CLS]} token of the annotator label is fed into a classification layer to predict the each annotator's label. As stated, we evaluate the predicted versus the true labels using cross entropy loss. Instances for which a particular annotator did not provide an annotation for the text were ignored when calculating the loss for that instance. 

\section{Results}

The task is evaluated using cross entropy loss (epsilon = $\ {1^{-12}}$) and F1 scores (average = `micro') of the hard and soft labels respectively. 

Our results are shown in \autoref{tab:all_results} and detailed results for all teams can be found in \citet{LeWiDi2023semeval}.Regarding F1 scores, the SVM model outperformed both the single-task BERT model, as well as the multi-task BERT model on the HS-Brexit, ConvAbuse, and MD datasets, findings which align with \citet{niemann-etal-2020-abusive}.The single-task BERT model outperformed both other architectures in the ArMIS dataset. 
For cross entropy scores, the single-task BERT model outperformed both other architectures across all datasets. Our model performs best on the ConvAbuse dataset, followed by the HS-Brexit dataset across both metrics. 

\section{Further Analyses}
\label{sec:further}

For deeper analysis, we used different methods depending on the specifics of the datasets.
For ConvAbuse and MD, 
we followed
\citet{davani-etal-2022-dealing} to deal with \emph{missing annotator labels} during the evaluation stage. Although not all annotators annotated every instance, our model still predicts labels for all annotators in the dataset
(e.g. predicting eight annotator labels in ConvAbuse for all instances that only have four true annotator labels as shown in \autoref{tab:kripps_score}). 
Essentially, predictions of  missing annotator labels might have negatively impacted the soft-label cross entropy comparisons by skewing distributions. This is especially the case for the MD dataset, where each instance was annotated by only five of the 891 annotators.

As such, we constrained the model to only predict labels for existing annotators of each instance, and reevaluated the soft-loss cross entropy and F1 scores. However, unlike \citet{davani-etal-2022-dealing}, our results degraded when constraining our model for both the ConvAbuse and MD datasets. Low performance in the MD dataset is further discussed in \autoref{sec: discussion-1}.   

We also investigated reasons for our poorest cross entropy scores, on the ArMIS dataset. We found that our model did not perform as well as expected in this scenario in which it should theoretically have performed well, i.e. with the annotators of the dataset self-identifying with a distinct ideological background (\emph{conservative}, \emph{moderate}, and \emph{liberal}) and no missing annotator labels. We found that this may have been the result of the model's architecture. Through testing with different batch sizes, we found that our model performed better when we added an extra hidden layer of 384 units to the existing 768-unit one. This would indicate that while the model was indeed learning, the original architecture with a single hidden layer was not sufficient to adequately disentangle enough information to make accurate predictions. 

\section{Discussion of Results \& Limitations} \label{sec: discussion}

\subsection{Performance evaluation}\label{sec: discussion-1}

We observe that our multi-task BERT model performed relatively well on the HS-Brexit and ConvAbuse datasets, but scored the lowest on the ArMIS and MD datasets. We expected to have limited success with the MD dataset, as multi-task models have an inherent issue with sparse data when combined with a  
high numbers of subtasks \citep{https://doi.org/10.48550/arxiv.1706.05098,9392366}. Not only does the MD-dataset contain more than 800 individual annotators, but the number of instances each annotator appeared in varies drastically (range $1-1988$, mean $=65.63\pm143.73$). Furthermore, not all annotators appear in all splits, with 91\% not represented in at least one of the training, development, and test sets. Individual annotator perspective modelling is therefore unfeasible for this dataset.

Such issues might be addressed through a clustering approach of labels. For example, \citealp{akhtar2019new, akhtar2020modeling, akhtar2021whose, dethlefs-etal-2014-cluster} have proposed clustering methods based on a variety of features, including demographic similarities, as well as using inter and intra-annotator disagreement and similarities in labelling behaviour. However, as stated by \citet{akhtar2021whose}, a limitation of their approach is that it is necessary to know the demographic and cultural background of annotators, which is information that is not available for the MD dataset. We plan to investigate clustering methods in our future work.

As previously stated, our model performed the worst on the ArMIS dataset. While some strategies to improve these scores were discussed in \autoref{sec:further}, we believe a possible explanation of these scores could be due to the size discrepancy of the datasets used in the original \citet{davani-etal-2022-dealing} paper compared to the size of the ArMIS dataset. The original multi-task model for example, used datasets with $\approx 30,000 - 60,000$ instances in each dataset. This shows that while a dataset may contain distinct annotator perspectives (e.g. evident in the ArMIS dataset both by self-declared ideological categories and a moderate inter-annotator disagreement of $0.524$), the multitask approach may not perform well on smaller datasets. 

Another possible reason might have been our loss-weighing strategy, which sums the individual cross entropy across subtasks. \citet{gong2019comparison} explain that such multi-task approaches, where the model loss is constituted by the sum loss of subtasks, can lead to degradation of performance. This is due to a possible conflict arising through contrasting losses between subtask, or conflicting gradient signals \citep{chen2018gradnorm, sener2018multi}. This aligns with our experience during training, where summed loss remained relatively stable, while individual loss across subtasks widely fluctuated. Furthermore, our soft-label cross entropy and F1 scores slowly improved over time in spite of the stable summed loss, indicating that some learning across subtasks was indeed taking place. 

\subsection{Capturing disagreement versus capturing perspectives} \label{sec: discussion-2}

The single-task BERT baseline model outperformed the multi-task model across all evaluative metrics.
\citet{cabitza-etal-2023-toward} explains that models trained through strong perspectivist approaches may be negatively impacted in terms of performance and evaluation metrics -- as the more nuance present in data (such as disaggregated annotations), the more difficult the data is to model. While our model exhibits these weaknesses, there are \emph{clear reasons} to use this approach; i.e. to model perspectives as an interim step in predicting disagreement. Our approach is successful in ways that would not be accounted for by simply predicting disagreement without this interim step.

Disagreement only shows that different perspectives are present in the dataset, but not the underlying reasons as to \emph{why} disagreement may occur, \emph{nor} the clashing perspectives present in the dataset.  Particularly for highly subjective tasks, modelling only the level of disagreement does not consider intersecting perspectives. In contrast, strong perspectivist approaches offer insights into the different opinions present amongst individuals or groups of annotators. Modelling perspectives does not erase these individual viewpoints. 
For example, research has shown that attributes such as gender \citep{waseem-hovy-2016-hateful}, or political activism status \citep{luo-etal-2020-detecting} of an annotator can elicit meaningful differences of opinions in a dataset. 

Furthermore, it is well documented that aggregation can harm minorities present in a dataset by limiting their opinion's influence \citep{prabhakaran-etal-2021-releasing}. \citet{gordon2022jury} explain that merely capturing disagreement can have a similar effect by presenting a simplified view of opposing perspectives in the data. This can be problematic, as without a nuanced understanding of which perspectives exist within a dataset, model predictions might not generalise well to end users' perspectives \citep{gordon2021disagreement}. 

Accurately predicting each annotator's perspective also captures their biases. However, bias is not an inherently negative trait. Though seldom explicitly stated, bias is an intrinsic attribute of annotators, datasets, and trained models \citep{gordon2022jury}. While (de)biasing models can lead to positive outcomes when attempting to make a model `unlearn' harmful social biases \citep{liang-etal-2020-towards, orgad2022debiasing, subramanian-etal-2021-evaluating}, \citet{devinney2022theories} assert that incorporating bias stemming from marginalised groups while training, can lead to models that amplify the voice of those minorities. For example, in cases with datasets dealing with gender-based violence \citep{cercas-curry-etal-2021-convabuse}, it might be preferable to capture and amplify the bias of the affected people. 

Combining approaches such as ours with the clustering approaches mentioned in \autoref{sec: discussion-1}, merits future research, especially since fully debiasing models seems improbable \citep{gordon2022jury}. As such, future research should attempt to utilise such multi-task models and strong perspectivist approaches when dealing with subjective tasks, in order to get a deeper understanding about why disagreement occurs.

\section{Conclusion}

We evaluated the performance of a multi-task model on predicting disagreement in four datasets, evaluated with both hard and soft labels, through F1 and cross-entropy loss respectively. Our model learned and predicted individual annotator perspectives for each instance.

Our model's findings did not outperform our single-task BERT baseline in terms of the shared tasks evaluation metric. This was due to model employing a strong perspectivist approach, which prioritised capturing individual perspectives present in the dataset, over high performance. 
We argue that a strong perspectivist approach is preferable to merely modelling disagreement as it allows to capture different opinions present in a dataset, and can be used to further amplify minority views. 

Evaluation metrics for this edition of Le-Wi-Di are geared towards measuring the overall levels of disagreement present in the datasets. 
However, if we wish to model stronger versions of perspectivism, we will need to develop new, more suitable metrics that can capture varying judgements in the kinds of different scenarios we have seen here. 

\section{Ethical considerations}

\paragraph{Reproducibility} 
We aim to maximise reproducability by making all data manipulation and modelling architecture aspects as explicit as possible in line with reproducibility principles \citep{belz-etal-2021-systematic}. The code used to produce this study's results can be found online in our team's \href{https://github.com/Ni-Vi/lewidi_2023}{GitHub repository}. 
\paragraph{Data manipulation and misrepresentation} Following concerns about possible mishandling of data in this study, an important point has to be made about the ConvAbuse dataset \citep{cercas-curry-etal-2021-convabuse}. As explained in \autoref{sec: data}, annotations ranged from $[-3, 1]$ with $1$ denoting no abuse, $0$ ambivalence, and the rest indicating severity of abuse.While analysing the dataset of this challenge, labels were aggregated to a binary depending on whether abuse was detected (labels -3,-1 in ConvAbuse), while the rest being annotated to no abuse detected. This transformation was necessary for the purposes of the shared task, as it has been shown that comparing scores between datasets with binary annotations and datasets with multiple labels can lead to incomparable results \citep{poletto2019annotating}. 

\paragraph{Abuse detection, and handling of sensitive opinions}  
There is also a larger conversation to be had about the use of a hard label in general, regarding issues such as bias, abuse, and such sensitive topics. Through a strong perspectivist approach, an annotator's viewpoint might also reflect the perspective of a minority population \citep{cabitza-etal-2023-toward}. It is important to be sensitive when dealing with such opinions without invalidating them, or minimising them through a distribution. In essence, if there is even a small amount of disagreement in whether an item is problematic or not, and the label should reflect that beyond a binary \citep{blodgett-2021-socio}. We leave it to future research to explain how exactly these multiple labels could be appropriately incorporated into a model architecture.  

\paragraph{Dual use of model}  
Finally, it is also important to explain that the model architecture proposed can also unfortunately be used for purposes beyond our original intention \citep{hovy2021five}.
While our model can be used towards furthering social justice aims through the amplification of minorities' perspectives, the model could also be used to manipulate perspectives from the dataset, in order to present stable results. 
For example, approaches have attempted to create perspective clusters that are more inclusive towards the data, \citep{subramanian-etal-2021-evaluating}. 
Unfortunately, these findings seem to result in bias mitigation \citep{shen-etal-2022-optimising,subramanian-etal-2021-fairness}, rather than bias erasure in both downstream tasks \citep{lalor-etal-2022-benchmarking}, as well as resulting word embeddings \citep{gonen-goldberg-2019-lipstick-pig}. As biases are unavoidable, we advocate boosting of under-respresented perspectives that might otherwise be lost. 
Such an approach would also make attempts to use this model to reproduce and platform problematic opinions transparent.

\section{Acknowledgements}
Gavin Abercrombie and Verena Rieser were supported by the EPSRC projects `Gender Bias in Conversational AI' (EP/T023767/1) and `Equally Safe Online (EP/W025493/1).
Tanvi Dinkar and Verena Rieser were supported by `AISEC: AI Secure and Explainable by Construction' (EP/T026952/1).

\bibliography{anthology,custom}
\bibliographystyle{acl_natbib}

\appendix
\section{Hyper-parameters for our experiments}

Both AraBERT and BERT consist of 12 encoder blocks, 12 attention heads, a hidden layer size of 768, while for the purposes of this study; maximum sequence length for both was set to 128.  Neural models learned through a fixed learning rate of $\ {1^{-8}}$, and a batch size of 64, over a maximum of 100 epochs. 
We used early stopping with a patience of 40 epochs while minimising the cross entropy scores obtained through the validation set.

\label{sec:appendix}

\end{document}